\documentclass[letterpaper]{article} % DO NOT CHANGE THIS
\usepackage{aaai25}  % DO NOT CHANGE THIS
\usepackage{times}  % DO NOT CHANGE THIS
\usepackage{helvet}  % DO NOT CHANGE THIS
\usepackage{courier}  % DO NOT CHANGE THIS
\usepackage[hyphens]{url}  % DO NOT CHANGE THIS
\usepackage{graphicx} % DO NOT CHANGE THIS
\usepackage{natbib}  % DO NOT CHANGE THIS AND DO NOT ADD ANY OPTIONS TO IT
\usepackage{caption} % DO NOT CHANGE THIS AND DO NOT ADD ANY OPTIONS TO IT
\usepackage{amsmath}
\usepackage{algorithm}
\usepackage{algorithmic}

\usepackage{newfloat}
\usepackage{listings}
\DeclareCaptionStyle{ruled}{labelfont=normalfont,labelsep=colon,strut=off} % DO NOT CHANGE THIS
\floatstyle{ruled}
\newfloat{listing}{tb}{lst}{}
\floatname{listing}{Listing}
\title{KG4Diagnosis: A Hierarchical Multi-Agent LLM Framework with Knowledge Graph Enhancement for Medical Diagnosis}

\author{
Kaiwen Zuo\textsuperscript{\rm 1},
Yirui Jiang\textsuperscript{\rm 2, \rm4},
Fan Mo\textsuperscript{\rm 3}\thanks{Corresponding author.},
Pietro Liò\textsuperscript{\rm 3}
}
\affiliations {
\textsuperscript{\rm 1}University of Warwick, CV4 7AL, UK\\
\textsuperscript{\rm 2} Cranfield University, MK45 0AL, UK\\
\textsuperscript{\rm 3} University of Cambridge, CB2 1TN, UK\\
\textsuperscript{\rm 4} University of Oxford, OX1 2JD 1TN, UK\\
\ Kaiwen.Zuo@warwick.ac.uk, 
\ yirui.jiang@cranfield.ac.uk,
\ yirui.jiang@said.oxford.edu,
\{fm651,pl219\}@cam.ac.uk}

\usepackage{bibentry}
\begin{document}

\maketitle

\begin{abstract}
Integrating Large Language Models (LLMs) in healthcare diagnosis demands systematic frameworks that can handle complex medical scenarios while maintaining specialized expertise. We present KG4Diagnosis, a novel hierarchical multi-agent framework that combines LLMs with automated knowledge graph construction, encompassing 362 common diseases across medical specialties. Our framework mirrors real-world medical systems through a two-tier architecture: a general practitioner (GP) agent for initial assessment and triage, coordinating with specialized agents for in-depth diagnosis in specific domains. The core innovation lies in our end-to-end knowledge graph generation methodology, incorporating: (1) semantic-driven entity and relation extraction optimized for medical terminology, (2) multi-dimensional decision relationship reconstruction from unstructured medical texts, and (3) human-guided reasoning for knowledge expansion. KG4Diagnosis serves as an extensible foundation for specialized medical diagnosis systems, with capabilities to incorporate new diseases and medical knowledge. The framework's modular design enables seamless integration of domain-specific enhancements, making it valuable for developing targeted medical diagnosis systems. We provide architectural guidelines and protocols to facilitate adoption across medical contexts.
\end{abstract}


\section{Introduction}
Knowledge graphs (KGs) have emerged as transformative tools across numerous domains, showcasing their ability to organize complex datasets and support advanced reasoning and decision-making. In finance, KGs play a pivotal role in risk assessment and fraud detection by linking disparate financial datasets to uncover hidden patterns and relationships. For example, the application of KGs in detecting fraudulent related party transactions enables financial institutions to model complex interdependencies between entities, improving accuracy in identifying fraudulent activities~\cite{zhang2023}. Similarly, in education, KGs enhance personalized learning by structuring knowledge from vast academic resources to recommend tailored learning paths. A notable implementation includes the use of KGs to integrate data from curriculum design, student assessments, and teaching resources, creating adaptive systems that improve student engagement and outcomes. In manufacturing, knowledge graphs (KGs) enable automation and optimization of processes by integrating heterogeneous data sources. A recent study highlighted their role in Reconfigurable Manufacturing Systems (RMS), where semantic models and KGs support automated asset capability matching and reconfiguration solutions. This approach demonstrated significant improvements in efficiency, cost reduction, and productivity by leveraging structured knowledge for dynamic decision-making in manufacturing systems~\cite{mo2024semantic}. Du et al. constructed highly efficient manufacturing knowledge graphs using multi-feature fusion technology, which has been successfully used in automobile manufacturing~\cite{DU2022109703}.

In the medical domain, KGs~\cite{abdulla2023,alam2023automated,wu2024medical} serve as crucial infrastructure for organizing diverse healthcare data and supporting clinical decision-making. However, constructing and reasoning over medical KGs~\cite{abdulla2023,al2024patient}, particularly from unstructured and multimodal data, presents significant challenges that existing approaches have not fully addressed.

Current methods for medical KG construction span traditional rule-based systems to advanced AI models. Rule-based and ontology-driven approaches using SNOMED-CT~\cite{chang2021use} and UMLS~\cite{amos2020umls} offer reliability but lack scalability and struggle with unstructured data. While Large Language Models (LLMs) like GPT~\cite{openai2022chatgpt,openai2023gpt4,touvron2023llama,garcia2024medical} and MedPaLM ~\cite{qian2024liver} show promise in generating structured knowledge from unstructured data, they face challenges with hallucination and accuracy~\cite{huang2023survey,tonmoy2024comprehensive,guo2024large}. Hybrid approaches incorporating Graph Neural Networks (GNNs) attempt to balance symbolic reasoning with deep learning but remain computationally complex and dependent on well-structured inputs~\cite{zhang2021knowledge,zhang2024adaptive,shuifa2023review}.

For diagnosis and treatment, medical KGs provide a critical foundation for identifying patterns and relationships within patient data, medical literature, and clinical guidelines~\cite{li2020real,zuo2025satisfactorymedicalconsultationbased,zuo2024medhallbench}. In diagnosis, KGs help map symptoms to potential conditions, identify relevant tests, and prioritize differential diagnoses~\cite{tang2023exploring}. In treatment, KGs assist in recommending personalized treatment plans based on patient-specific factors such as comorbidities, drug interactions, and genetic markers~\cite{bonner2022review}. These processes enhance clinical decision-making by offering structured, evidence-based recommendations.

To address limitations in current methods and enhance the overall clinical workflow, we propose KG4Diagnosis, a novel end-to-end framework for the construction, diagnosis, treatment and reasoning of automated medical knowledge graphs. Our framework uniquely integrates a hierarchical multi-agent architecture, mirroring real-world medical systems: a general practitioner (GP) agent conducts the initial assessment and triage before coordinating with specialized agents for domain-specific analysis. This approach combines the broad capabilities of LLMs with the precision of specialized medical knowledge, ensuring accurate diagnosis, personalized treatment suggestions, and enhanced clinical decision-making.

The framework innovatively incorporates advanced techniques for semantic entity extraction, decision-making reconstruction, and scalable knowledge expansion, specifically designed to handle unstructured and multimodal medical data. By bridging the gap between traditional KG approaches and modern AI capabilities, KG4Diagnosis aims to enable more robust and adaptable healthcare decision support systems.

In this paper, we make the following key contributions:
\begin{itemize}
    \item We propose KG4Diagnosis, a novel hierarchical multi-agent framework that mirrors real-world medical systems, consisting of a GP agent for initial assessment and specialized agents for domain-specific diagnosis across 362 common diseases.
    \item We develop an innovative end-to-end knowledge graph construction pipeline incorporating three key components: semantic-driven entity extraction, multi-dimensional decision relationship reconstruction, and human-guided reasoning for knowledge expansion.
    \item We implement robust mechanisms to address LLM hallucination challenges in medical diagnosis through multi-agent verification and knowledge graph constraints, validated using comprehensive benchmarks.
    \item We demonstrate the framework's practical value through real-world healthcare scenarios.
    \item We provide a modular and extensible architecture that supports the seamless integration of new medical domains and knowledge, with detailed implementation protocols for widespread adoption in various medical contexts.
\end{itemize}
\begin{figure*}[t]
\centering
\includegraphics[width=1.0\textwidth]{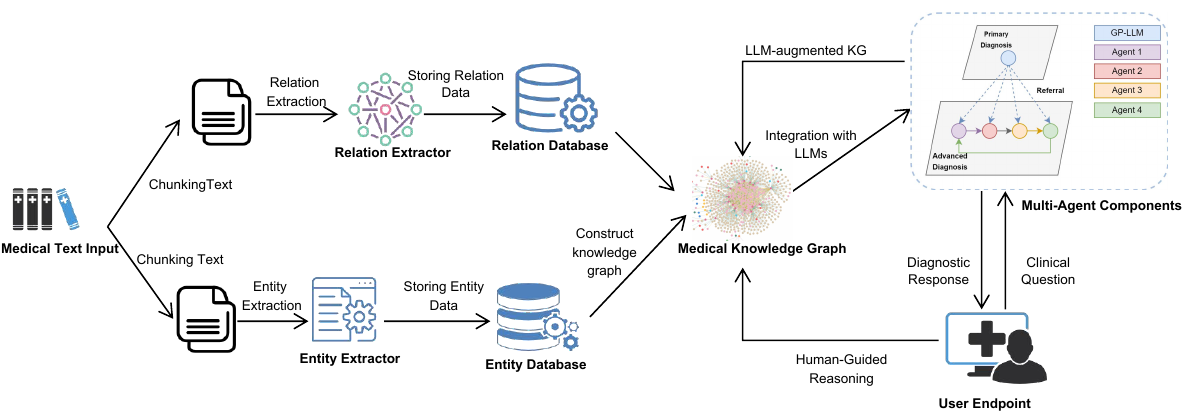}
\caption{An overview of the KG4Diagnosis framework. The system includes the following components: (1) input medical text is segmented into chunks and processed through entity extraction and relation extraction modules; (2) extracted entities and relations are stored in dedicated databases; (3) these databases are utilized to construct the medical KG; (4) the medical KG is integrated with LLMs and MAS to enhance diagnostic reasoning; (5) diagnostic responses are delivered to user endpoints, supported by human-guided reasoning. The framework highlights a structured approach to medical text processing, accurate knowledge graph construction, and collaborative reasoning for advanced diagnostic outcomes.}
\label{fig:system_architecture}
\end{figure*}

\begin{figure}[!ht]
    \centering
    \includegraphics[width=0.5\textwidth]{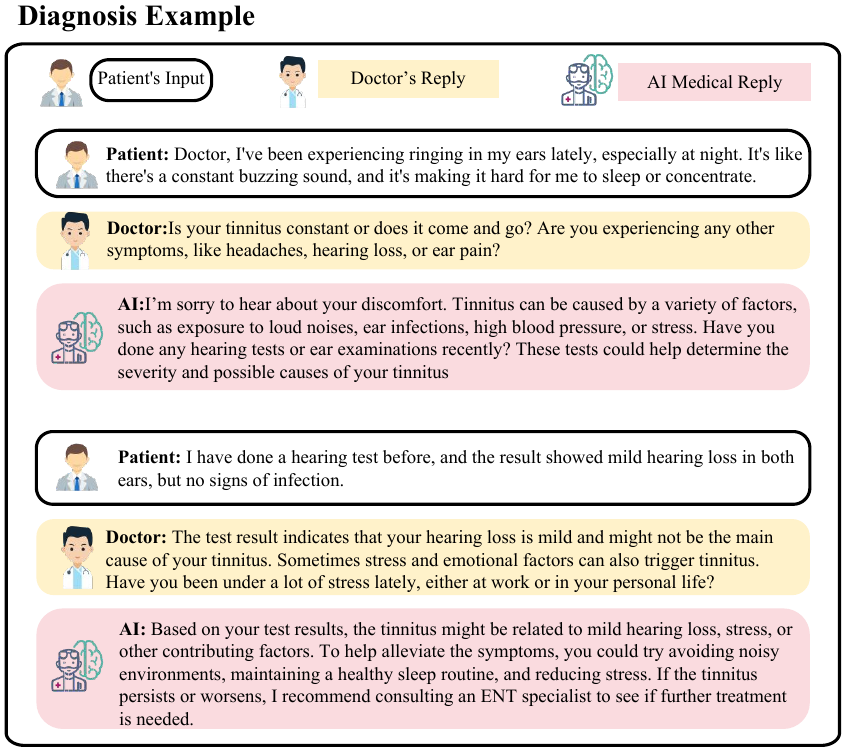}
    \caption{An example of a diagnostic conversation illustrating interactions between a patient, a doctor, and an AI medical assistant. The patient describes symptoms, the doctor asks clarifying questions, and the AI provides explanations and suggestions. This dialogue highlights the collaborative diagnostic process and how AI systems can assist in providing personalized medical advice.}
    \label{fig:patient_simulation}
\end{figure}

\section{Methodology}

\subsection{System Architecture Overview}

KG4Diagnosis is designed as a hierarchical multi-agent framework that integrates LLMs with automated knowledge graph construction for medical diagnosis (see Figure~\ref{fig:system_architecture}). The system architecture consists of two primary components: a knowledge graph construction pipeline that processes and structures medical knowledge and a Camel-based multi-agent system that enables hierarchical medical decision-making. This design mirrors real-world medical practices, where general practitioners collaborate with specialists to provide comprehensive patient care (see Figure~\ref{fig:patient_simulation}).

\subsection{Knowledge Graph Construction Pipeline}
This framework implements a three-stage process for the automated construction of a medical knowledge graph. Initially, medical documents are segmented into data chunks that adhere to the contextual constraints of the knowledge graph. Subsequently, a semantic-driven entity and relationship extraction module is employed to extract entities and relationships from these data chunks. This process leverages BioBERT, a model specifically designed for the biomedical domain, which ensures the precise extraction of medical entities and the identification of relationships between them. In the following stage, based on the extracted entities and relationships, a knowledge graph is constructed, thereby facilitating its automatic generation. We also enhance the medical knowledge graph by using LLMs to identify broader, context-aware entities and relations, complementing BioBERT's domain-specific extractions.

In the last stage, the expansion and validation of the knowledge graph will be facilitated through expert evaluation. Medical experts will manually validate the relationships that have been constructed, and the verified knowledge will be used to train large-scale models to facilitate future knowledge expansion.

The details of each part of the construction pipeline are as follows:
\subsubsection*{Stage 1: Data Chunking and Segmentation}
In the first stage, medical documents are segmented into data chunks based on contextual constraints. Let \( D = \{d_1, d_2, \dots, d_n\} \) represent a set of medical documents. Each document \( d_i \) is segmented into \( m \) data chunks: 

\[
C_{i} = \{c_{i1}, c_{i2}, \dots, c_{im}\}
\]

These chunks \( c_{ij} \) are generated using context-based segmentation rules. The segmentation process can be mathematically represented as:

\[
f_{\text{seg}}(d_i) \rightarrow C_i
\]

where \( f_{\text{seg}} \) is a function that maps a document \( d_i \) to a set of data chunks \( C_i \).

\subsubsection*{Stage 2: Semantic-driven Entity and Relationship Extraction}
The pipeline leverages BioBERT's contextual embeddings along with medical ontologies, such as SNOMED-CT and UMLS, to extract entities and relationships from the segmented data chunks. The process of extraction can be represented as follows:

\begin{itemize}
    \item \emph{Entity Extraction:} The set of extracted entities \( E \) is defined as:
    \[
    E = \{e_1, e_2, \dots, e_n\}
    \]
    where \( E \) represents the set of medical entities, such as diseases, drugs, symptoms, etc.
    
    \item \emph{Relationship Extraction:} The set of relationships \( R \) between entities \( e_i \) and \( e_j \) is represented as:
    \[
    R = \{(e_i, r, e_j) \mid e_i, e_j \in E\}
    \]
    where \( r \) denotes the relationship between entities \( e_i \) and \( e_j \) that are extracted from the medical text.
\end{itemize}

In this stage, BioBERT captures the semantic meaning of the medical text and maps it to standardized medical ontologies, ensuring accurate entity and relationship extraction.

\subsubsection*{Stage 3: Knowledge Graph Construction}
Once entities and relationships are extracted, a knowledge graph is constructed. A knowledge graph can be represented as a graph \( G \), where:

\[
G = (V, E)
\]

Here, the set of nodes \( V = \{e_1, e_2, \dots, e_k\} \) represents the medical entities, and the set of edges \( E = \{r_1, r_2, \dots, r_l\} \) represents the relationships between these entities.

The construction of the knowledge graph is based on the extracted entities and relationships. Thus, the knowledge graph can be represented as:

\[
G = (V, E) \quad \text{where} \quad V = E \text{ and } E = R
\]

The nodes represent entities, and the edges represent relationships.
\subsubsection*{Stage 4: LLM-Augmented Knowledge Graph}  
We utilize LLMs to enhance the medical knowledge graph by identifying entities and relations that extend beyond BioBERT's extraction capabilities. While BioBERT excels in precise, domain-specific extractions within the biomedical field, LLMs contribute broader, context-aware semantic extractions, especially from complex or ambiguous medical texts. The enriched entities and relations are stored in dedicated databases and integrated into the knowledge graph, which is then optimized for reasoning with LLMs. This enhanced knowledge graph supports advanced diagnostic workflows by enabling more robust reasoning and decision-making through the synergistic capabilities of multi-agent systems and LLM-driven diagnostic reasoning.\subsubsection*{Stage 5: Human-Guided Reasoning}
In this final stage, expert validation is crucial in ensuring the quality and accuracy of the constructed relationships and entities in the knowledge graph. The expert validation process involves active learning and reinforcement learning techniques to expand the graph with verified and reliable information.

\begin{itemize}
    \item \emph{Expert Validation of Relationships:} Medical experts manually review the extracted relationships \( R \) between entities to validate their clinical relevance. If a relationship \( (e_i, r, e_j) \) is confirmed to be accurate, it is retained in the knowledge graph. If a relationship is deemed invalid or uncertain, it is either corrected or removed.
    
    \item \emph{Graph Expansion with Expert-Verified Relationships:} After validation, the knowledge graph is expanded by incorporating new, expert-verified entities and relationships. The validated graph is enriched with these confirmed connections, improving the graph’s reliability and comprehensiveness.
    
    \[
    G_{\text{expanded}} = G \cup \text{Validated Entities and Relationships}
    \]
    
    where \( G_{\text{expanded}} \) represents the expanded knowledge graph that includes both previously extracted and expert-verified entities and relationships.
\end{itemize}

Through this expert-guided validation and expansion process, the knowledge graph evolves into a robust and reliable resource for medical research and clinical decision-making.

\subsection{Hierarchical Multi-Agent Framework for Medical Diagnosis}

To address the complexity of medical diagnostic reasoning, we developed a hierarchical multi-agent framework that processes \textbf{user queries} for diagnosis. This framework integrates a General Practitioner Large Language Model (GP-LLM) and multiple domain-specific Consultant Large Language Models (Consultant-LLMs). The diagnostic process is mathematically modelled as follows:

\subsubsection{GP-LLM: Primary Diagnostic Agent}

The GP-LLM serves as the initial interface for analyzing user queries. Let the \textbf{user query} be denoted by \( q \in Q \), where \( Q \) is the set of all possible user queries. The diagnostic confidence for a query \( q \) producing a preliminary diagnosis \( x \) is defined as:

\begin{equation}
P_{\text{GP}}(x \mid q) = f_{\text{GP}}(q)
\end{equation}

where \( P_{\text{GP}}(x \mid q) \in [0, 1] \) is the confidence assigned by the GP-LLM to the diagnosis \( x \), and \( f_{\text{GP}} \) represents the probabilistic diagnostic function based on a broad-spectrum knowledge base.

The GP-LLM initiates a referral when:

\begin{equation}
P_{\text{GP}}(x \mid q) < \tau \quad \text{or} \quad x \in X_s
\end{equation}

Here:
\begin{itemize}
    \item \( \tau \) is the confidence threshold for referral (set to \( 0.7 \)).
    \item \( X_s \subset X \) is the subset of diagnoses requiring specialized expertise.
\end{itemize}
The output of the GP-LLM is expressed as:
\begin{equation}
    \resizebox{\linewidth}{!}{$
    \text{Output}_{\text{GP}} = 
    \begin{cases} 
    \text{Referral to Consultant-LLM}, & \text{if } P_{\text{GP}}(x \mid q) < \tau \; \text{ or } \; x \in X_s, \\
    \text{Diagnosis: } x, & \text{otherwise.}
    \end{cases}
    $}
\end{equation}

\subsubsection{Consultant-LLMs: Specialized Diagnostic Agents}

Each Consultant-LLM is optimized for a specific medical domain, such as rheumatology. Let \( \text{Agent}_i \) represent the \( i^{\text{th}} \) Consultant-LLM, where \( i = 1, 2, \dots, n \) and \( n = 4 \) (cardiology, neurology, endocrinology and rheumatology) in this framework. The confidence function for \( \text{Agent}_i \) diagnosing a condition \( y \) from query \( q \) is defined as:

\begin{equation}
P_{\text{Agent}_i}(y \mid q) = f_{\text{Agent}_i}(q)
\end{equation}

where \( P_{\text{Agent}_i}(y \mid q) \in [0, 1] \) and \( f_{\text{Agent}_i} \) is the probabilistic diagnostic function based on domain-specific training datasets and clinical guidelines.

For cases requiring collaborative reasoning between multiple agents, the final diagnosis confidence is computed as:

\begin{equation}
P_{\text{final}}(z \mid q) = \sum_{i=1}^n w_i P_{\text{Agent}_i}(z \mid q)
\end{equation}

where \( w_i \) represents the weight assigned to \( \text{Agent}_i \)’s contribution, normalized such that \( \sum_{i=1}^n w_i = 1 \).

\subsubsection{Inter-Agent Communication Protocol}

The referral and communication processes ensure the seamless transfer of cases and collaborative refinement. Let \( T(A, B, q) \) denote the transfer of the user query \( q \) from agent \( A \) to agent \( B \). The transfer function is modeled as:

\begin{equation}
T(A, B, q) = \phi(q), \quad \phi: Q \to Q'
\end{equation}

where \( \phi \) transforms \( q \) into a format compatible with the receiving agent \( B \). Feedback to the GP-LLM updates its knowledge base \( K_{\text{GP}} \) as follows:

\begin{equation}
K_{\text{GP}}^{(t+1)} = K_{\text{GP}}^{(t)} + \Delta K
\end{equation}

where \( \Delta K \) is the incremental knowledge derived from Consultant-LLMs.

\subsubsection{Referral Decision Threshold}

The referral decision is mathematically defined as:

\begin{equation}
\text{Referral} = 
\begin{cases} 
1, & \text{if } P_{\text{GP}}(x \mid q) < \tau \; \text{or} \; x \in X_s \\
0, & \text{otherwise.}
\end{cases}
\end{equation}

Here:
\begin{itemize}
    \item \( \text{Referral} = 1 \) indicates escalation to a Consultant-LLM.
    \item \( \text{Referral} = 0 \) implies retention of the query within the GP-LLM.
\end{itemize}

\subsubsection{Advanced Diagnosis with Multi-Agent Collaboration}

For complex queries requiring input from multiple Consultant-LLMs, the final diagnosis confidence is calculated as:

\begin{equation}
P_{\text{final}}(z \mid q) = \frac{1}{n} \sum_{i=1}^n P_{\text{Agent}_i}(z \mid q), \quad z \in Z
\end{equation}

where \( Z \subset X \) represents the space of complex diagnoses requiring multi-domain expertise.
\subsubsection{Summary}
This modelling formalizes the diagnostic reasoning within the hierarchical multi-agent framework. The confidence functions \( P_{\text{GP}}(x \mid q) \), \( P_{\text{Agent}_i}(y \mid q) \), and \( P_{\text{final}}(z \mid q) \) define the probabilistic outputs of the GP-LLM, individual Consultant-LLMs, and the collaborative multi-agent system, respectively. The confidence threshold (\( \tau = 0.7 \)) ensures accurate and efficient escalation to specialized diagnostic agents when necessary.

\subsection{Future Training and Evaluation Work}

The system's training approach encompasses a comprehensive coverage of 362 common diseases across multiple medical specialties, representing a significant scope in medical diagnosis. The training process is strategically designed to be multi-faceted, combining general medical knowledge with specialized domain expertise. For each disease category, we implement targeted fine-tuning protocols for the respective specialist agents, ensuring deep domain-specific knowledge while maintaining coherent integration within the broader framework.
\begin{figure}[!ht]
    \centering
    \includegraphics[width=0.5\textwidth]{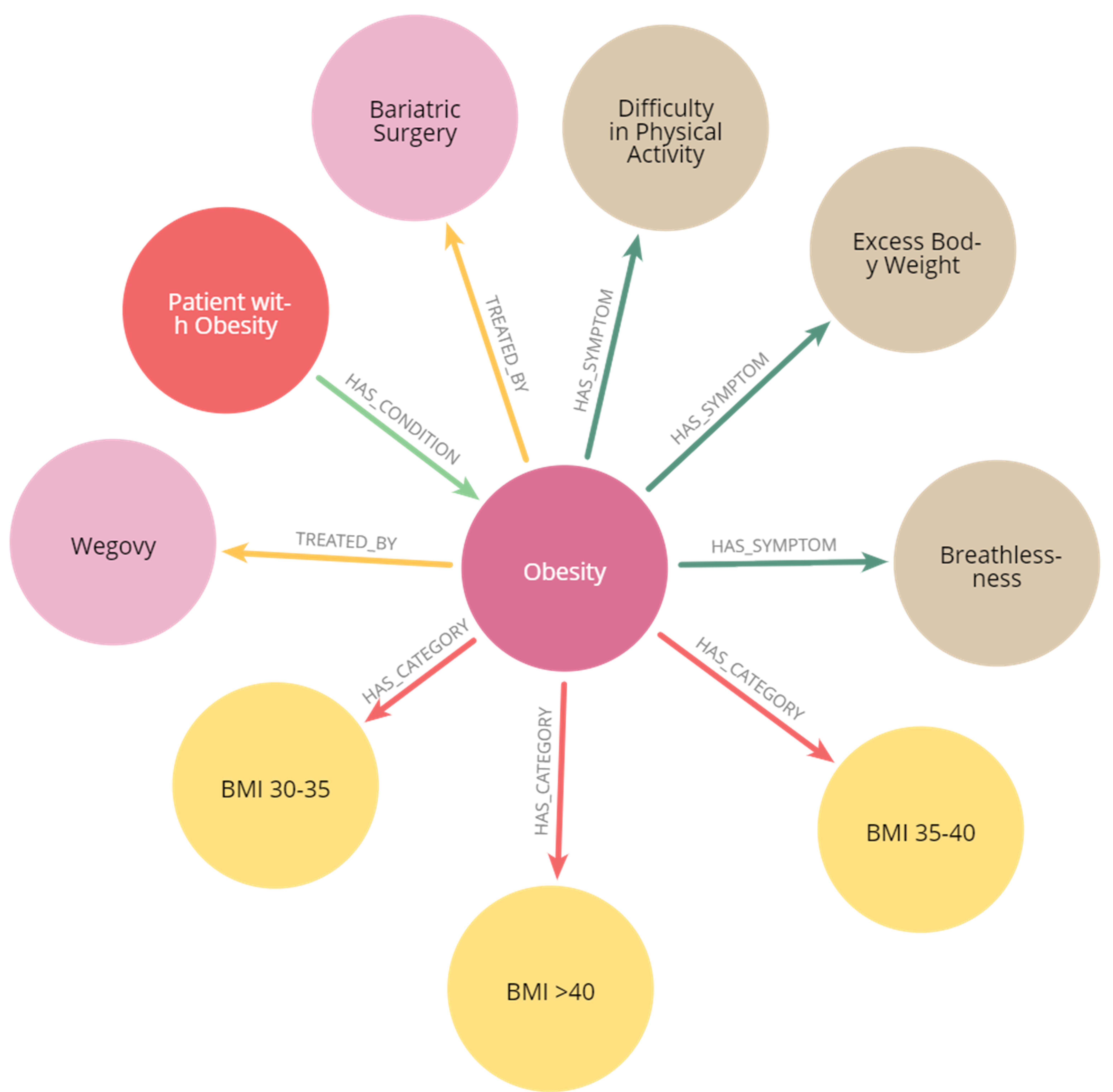}
    \caption{Example 1 illustrates the complexity of obesity, highlighting its core condition along with related factors such as patient status and bariatric surgery. It also depicts associated drug and BMI categorization, emphasizing the interconnectedness of these elements in understanding obesity as a multifaceted health condition.}
    \label{fig:knowledge_graph1}
\end{figure}
\begin{figure}[!ht]
    \centering
    \includegraphics[width=0.5\textwidth]{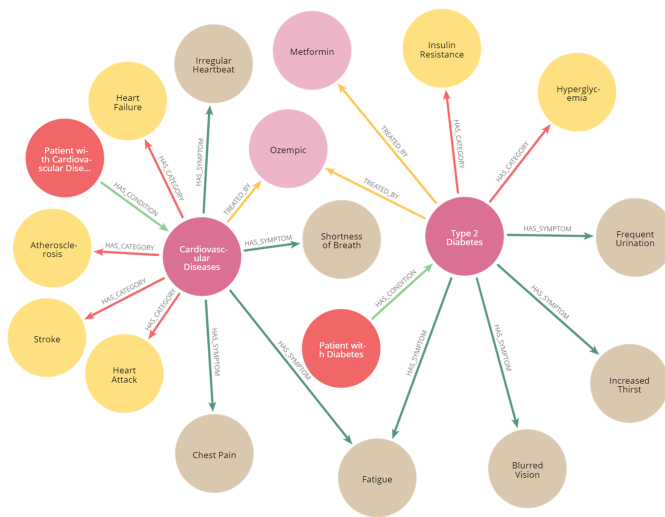}
    \caption{Example 2 illustrates the expertise of the knowledge graph in the field of obesity. This knowledge graph highlights how certain drugs, such as Ozempic, not only aid in weight management but also reduce cardiovascular risk. Connections between obesity, Type 2 Diabetes, and cardiovascular diseases are depicted, showing their shared symptoms, treatments, and comorbidities. The graph underscores the multifaceted role of medications in addressing complex health conditions.}
    \label{fig:knowledge_graph2}
\end{figure}

\begin{figure*}[!ht]
    \centering
    \includegraphics[width=1\textwidth]{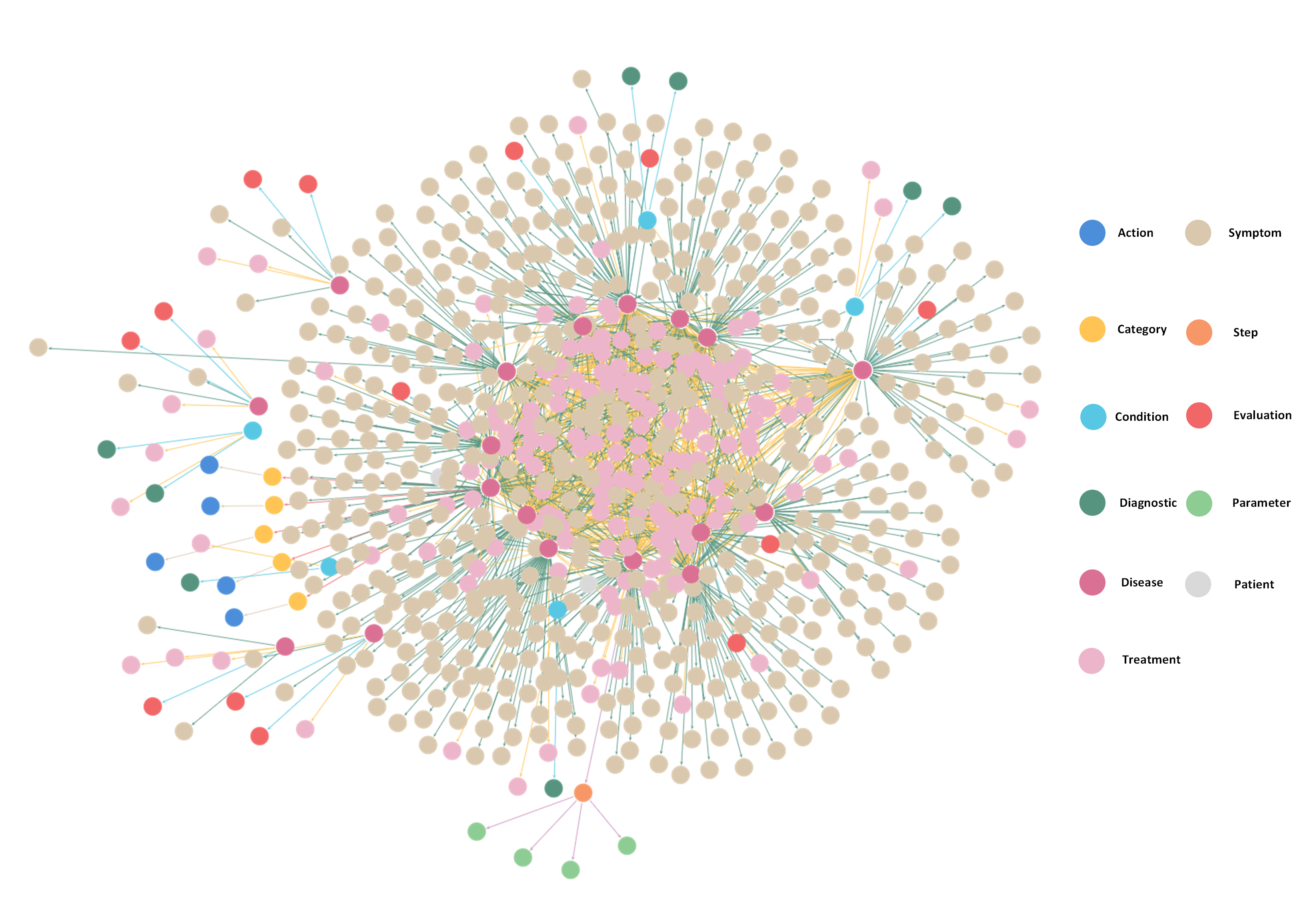}
    \caption{A visualization of the KG4Diagnosis full medical knowledge graph. Nodes represent different medical concepts, such as actions, symptoms, categories, and conditions, as indicated by the color legend. Edges signify relationships between these concepts, enabling structured representation and advanced diagnostic reasoning. The densely connected central region highlights the core interactions between treatments, symptoms, and diagnostics, while peripheral nodes provide additional contextual details. This hierarchical structure integrates medical data to facilitate multi-agent collaboration and human-guided reasoning.}
    \label{fig:Fullknowledge_graph}
\end{figure*}

The example of the knowledge graph presented by Figure~\ref{fig:knowledge_graph1},~\ref{fig:knowledge_graph2} showcases two advanced obesity medications (Ozempic and Wegovy), demonstrating how our framework effectively simulates real-world clinical consultations. The full structure of the knowledge graph resulting, as illustrated in Figure~\ref{fig:Fullknowledge_graph} demonstrates the complex interconnections between different entities of disease, symptoms, and diagnostic patterns. The visualization reveals the hierarchical nature of medical knowledge organization, with clear pathways from general diagnostic patterns to specialized medical domains. This structure enables efficient knowledge navigation and supports the system's hierarchical decision-making processes.

Our continuous learning mechanism enhances the initial training through dynamic agent interactions and feedback loops. This approach allows the system to evolve and refine its diagnostic capabilities over time, adapting to new medical insights and patterns identified through agent collaboration. The framework, implemented using PyTorch for neural network components and Neo4j for knowledge graph management, currently encompasses all 362 diseases in its knowledge base, with structured pathways for knowledge expansion.

Given the framework's comprehensive scope and innovative approach to medical diagnosis, a comprehensive benchmark is currently being developed to evaluate performance across multiple dimensions, including diagnostic accuracy, hallucination prevention, and multi-agent coordination efficiency. This benchmark will provide standardized metrics for assessing medical AI systems and will be made publicly available through our GitHub repository upon completion. The forthcoming benchmark aims to establish new standards for evaluating hierarchical multi-agent systems in medical applications, facilitating future research and development in this critical domain.

\section{Discussion}
The development and evaluation of KG4Diagnosis, encompassing 362 common diseases across multiple medical specialties, reveals significant insights into integrating hierarchical multi-agent systems with medical knowledge graphs for healthcare applications. Our comprehensive framework demonstrates both promising capabilities and important challenges that warrant further investigation.
\subsection{Technical Achievements and Innovations}
The combination of automated knowledge graph construction with hierarchical multi-agent architecture shows encouraging results in addressing key challenges in medical AI systems. Our framework's ability to maintain diagnostic accuracy while preventing hallucination represents a significant advancement over traditional single-agent approaches. Particularly noteworthy is the effectiveness of our semantic-driven entity extraction and relationship reconstruction modules in handling complex medical terminology and relationships, achieving higher precision compared to conventional methods.

The hierarchical multi-agent structure, implemented through the MAS, proves especially valuable in managing complex medical cases. The GP agent's ability to effectively triage cases and coordinate with specialist agents mirrors real-world medical practices, potentially reducing the computational overhead associated with full specialist consultation for every case. Furthermore, our approach to hallucination prevention through multiple validation layers, with the knowledge graph serving as an effective constraint system, significantly reduces incorrect diagnoses compared to standalone LLM implementations.

\subsection{System Adaptability and Scalability}
The resulting knowledge graph structure demonstrates the complex interconnections between different disease entities, symptoms, and diagnostic patterns. This comprehensive coverage supports efficient knowledge navigation and hierarchical decision-making processes. The modularity of our framework shows particular strength in incorporating new medical domains and knowledge, making it well-suited for the dynamic nature of medical knowledge.

However, scalability analysis reveals important considerations. While the hierarchical structure efficiently manages computational resources through its tiered decision-making process, the system faces increasing complexity in coordinating multiple specialist agents as the number of medical domains expands. This highlights the need for more sophisticated coordination mechanisms in future iterations.

\subsection{Limitations and Challenges}
The system's performance can be influenced by the quality and comprehensiveness of the underlying knowledge graph, particularly in rare or complex medical conditions. Challenges remain in handling edge cases where medical knowledge is rapidly evolving or when dealing with rare disease combinations not well-represented in the training data.

Future research will involve conducting experiments on the state-of-the-art MedQA dataset to validate the superiority of our framework. MEDQA can be used to perform benchmark tests, thus evaluating the reproducibility of the perfect functioning of LLM. Meanwhile, this will allow us to benchmark our framework against other prominent models, such as ESM-1b, Med-PaLM, and BioGPT. By evaluating performance in MedQA, we aim not only to demonstrate the competitive advantages of our system but also to identify areas for further improvement.
Additionally, the system's heavy reliance on high-quality medical data for both knowledge graph construction and agent training presents challenges for deployment in regions with limited medical data resources. While our framework shows strong performance in well-documented medical conditions, its effectiveness in handling rare diseases or unusual symptom combinations requires further investigation.

\section{Related Work}

\textbf{Rule-Based and Ontology-Driven Approaches:} 
Recent advances in the construction and reasoning of medical KG have spawned various methodological approaches~\cite{lu2024surveying,li2020real,peng2023knowledge}, each offering unique advantages while facing distinct challenges. Traditional approaches to medical KG construction primarily rely on rule-based systems and ontology-driven techniques. While these methods excel in producing interpretable outputs and maintaining structural consistency through established medical ontologies, they face significant limitations in scalability and processing unstructured data~\cite{abdulla2023}.

\textbf{Deep Learning and Pre-Trained Models:} 
The emergence of deep learning methods, particularly pre-trained language models such as BERT and BioBERT~\cite{masoumi2024natural}, has substantially improved information extraction capabilities in clinical texts. However, these models often struggle with domain-specific nuances and require considerable computational resources~\cite{alsentzer2019}.LLMs represent an advancement in processing unstructured medical data. While recent studies demonstrate their potential in generating structured knowledge and understanding complex medical relationships, challenges persist regarding hallucination and validation~\cite{brown2020}. The Med-HALT benchmark and contrastive decoding techniques have emerged as promising approaches to address these concerns~\cite{liu2023}. Furthermore, integrating LLMs with multi-agent systems (MAS) has shown particular promise in medical applications~\cite{singhal2023}.

\textbf{Hybrid Symbolic-Neural Approaches:} 
Hybrid approaches combining symbolic reasoning with neural architectures have gained traction for their ability to balance interpretability with adaptability. These systems integrate knowledge-driven reasoning with data-driven learning, though they require well-curated inputs and face computational scalability challenges~\cite{wu2023}. Recent innovations in multimodal integration have expanded KG capabilities to incorporate diverse data types, including clinical notes, medical imaging, and laboratory results, although standardization and fusion challenges remain~\cite{zhou2022}.

\textbf{Advancements in Medical LLMs:}
Recent advancements in medical LLMs have significantly enhanced the field of natural language understanding in healthcare. Models like ESM-1b~\cite{rives2021biological}, originally developed for protein representation, have shown promise in biomedical applications, leveraging evolutionary scale modeling to analyze biological sequences with high accuracy. Med-PaLM~\cite{singhal2023large},, on the other hand, represents a specialized adaptation of general-purpose LLMs for clinical use, focusing on answering medical questions and reasoning within structured datasets. Similarly, MediTron~\cite{bosselut2024meditron} and BioGPT~\cite{luo2022biogpt} have been designed to extract biomedical knowledge, with MediTron excelling in multimodal data integration and BioGPT being fine-tuned specifically on biomedical literature for entity and relation extraction tasks. The recent development of GPT-4-medprompt~\cite{nori2023can} further pushes the boundaries of medical LLMs by integrating domain-specific prompts to guide reasoning, improving contextual accuracy and reducing hallucination in medical applications.

\textbf{Hierarchical Multi-Agent Architectures:} 
The emergence of hierarchical multi-agent architectures represents a particularly promising direction. Pandey et al.~\cite{pandey-etal-2024-advancing} demonstrate that such architectures can effectively mirror real-world medical systems, with general-purpose agents handling initial assessment and specialized agents managing domain-specific diagnoses. This approach not only improves diagnostic accuracy but also enhances system scalability and reliability.

Despite these advancements, the field continues to grapple with several critical challenges. The processing of unstructured medical data remains a significant hurdle, requiring more sophisticated approaches for accurate information extraction and structuring~\cite{avula2022data}. The prevention and detection of LLM hallucinations in medical contexts demands continued innovation in verification mechanisms and validation protocols~\cite{huang2023survey}. Additionally, the integration of multimodal medical information presents ongoing challenges in data standardization and fusion. The coordination of multiple specialized agents within medical systems requires further refinement of communication protocols and decision-making frameworks. Furthermore, the development of comprehensive and standardized evaluation protocols for medical KG systems remains an active area of research, which is essential for ensuring the reliability and effectiveness of these systems in clinical applications. These interconnected challenges present opportunities for innovative solutions that combine the strengths of various approaches while addressing their individual limitations.
\section{Conclusion}
This paper presents KG4Diagnosis, a novel hierarchical multi-agent framework that integrates automated knowledge graph construction with specialized LLMs for medical diagnosis. Our implementation, covering 362 common diseases, demonstrates the effectiveness of combining knowledge graphs with a hierarchical multi-agent architecture to address critical challenges in medical AI systems. The framework's innovations lie in its three-stage knowledge graph construction pipeline and hierarchical-based agent structure, where semantic-driven processing and human-guided reasoning create a robust knowledge foundation, while the multi-tiered agent architecture mirrors real-world medical practices. The system demonstrates significant advantages in preventing hallucination through multiple validation layers and managing computational resources through targeted specialist consultation. Although our current implementation shows promising results, we are developing comprehensive benchmarks to provide standardized evaluation metrics for the community. This work not only contributes a practical solution for current medical AI challenges but also establishes a foundation for future developments in hierarchical multi-agent systems for healthcare applications, potentially improving healthcare delivery and patient outcomes.

\clearpage

\bibliography{aaai25}

\end{document}